\documentclass[10pt,twocolumn,letterpaper]{article}

\usepackage[pagenumbers]{cvpr} 

\usepackage{graphicx} 
\usepackage[utf8]{inputenc}
\usepackage[T1]{fontenc}
\usepackage{placeins}
\usepackage{amsmath}
\usepackage{amsfonts}
\usepackage{amssymb}
\usepackage{graphicx}
\newsavebox\CBox 
\usepackage{booktabs}
\usepackage{subcaption}
\usepackage{tabularx}
\usepackage{printlen}
\usepackage{stfloats}
\usepackage{enumitem}
\usepackage{adjustbox}
\usepackage{colortbl}
\usepackage{xcolor}
\usepackage{array}
\usepackage{fancyhdr}

\usepackage[accsupp]{axessibility}

\definecolor{cvprblue}{rgb}{0.21,0.49,0.74}
\usepackage[pagebackref,breaklinks,colorlinks,allcolors=cvprblue]{hyperref}

\def\paperID{V4A-26} 
\def\confName{CVPRW}
\def\confYear{2025}

\title{Weakly Supervised Panoptic Segmentation for Defect-Based Grading of\\Fresh Produce}

\author{
Manuel Knott$^{1,2,3}$
~~
Divinefavour Odion$^{4}$
~~
Sameer Sontakke$^{5}$
\\
Anup Karwa$^{5}$
~~
Thijs Defraeye$^{1}$
\\\\
$^{1}$\,Empa, Swiss Federal Laboratories for Materials Science and Technology, St. Gallen, Switzerland
\\
$^{2}$\,Swiss Data Science Center, ETH Zurich and EPFL, Switzerland
\\
$^{3}$\,Institute for Machine Learning, Department of Computer Science, ETH Zurich, Switzerland
\\
$^{4}$\,Constructor University, Bremen, Germany
\\
$^{5}$\,Innoterra BioScience Private Limited, Mumbai, India
}

\definecolor{LightBlue}{RGB}{198,221,236}

\definecolor{Yellow}{RGB}{255, 255, 0}
\definecolor{Orange}{RGB}{255, 165, 0}
\definecolor{Red}{RGB}{255, 0, 0}
\definecolor{Purple}{RGB}{128, 0, 128}
\definecolor{Pink}{RGB}{255, 192, 203}
\definecolor{Teal}{RGB}{0, 128, 128}

\newcommand*{\simsym}{\mathord\sim}

\begin{document}
\maketitle

\begingroup
\renewcommand{\thefootnote}{} 
\footnotetext{\hspace{-.5cm}Accepted as a paper to the \textit{6th International Workshop on Agriculture-Vision: Challenges \& Opportunities for Computer Vision in Agriculture} in conjunction with IEEE/CVF CVPR 2025.}
\addtocounter{footnote}{-1} 
\endgroup

\emergencystretch 3em
\begin{abstract}
Visual inspection for defect grading in agricultural supply chains is crucial but traditionally labor-intensive and error-prone. Automated computer vision methods typically require extensively annotated datasets, which are often unavailable in decentralized supply chains. We address this challenge by evaluating the Segment Anything Model (SAM) to generate dense panoptic segmentation masks from sparse annotations. These dense predictions are then used to train a supervised panoptic segmentation model. Focusing on banana surface defects (bruises and scars), we validate our approach using 476 field images annotated with 1440 defects. While SAM-generated masks generally align with human annotations, substantially reducing annotation effort, we explicitly identify failure cases associated with specific defect sizes and shapes. Despite these limitations, our approach offers practical estimates of defect number and relative size from panoptic masks, underscoring the potential and current boundaries of foundation models for defect quantification in low-data agricultural scenarios.
GitHub: \href{https://github.com/manuelknott/banana-defect-segmentation}{https://github.com/manuelknott/banana-defect-segmentation}

\end{abstract}

\section{Introduction}

Visual inspection of agricultural products plays a crucial role in postharvest supply chains. Particularly, fresh produce is often graded based on visible blemishes and defects. 
Traditionally, this process relies on manual inspection—an approach that is labor-intensive, time-consuming, and prone to inconsistency and human error. Additionally, inspectors require specialized training to recognize various types of defects, assess their size, and determine their severity. 
For those reasons, machine learning-based computer vision models have become increasingly popular for automated assessment, increasing the speed and consistency of grading \citep{yuvaraj_implementation_2023, barthwal_new_2024}.
These deep learning models traditionally require a considerably large and well-annotated training dataset to work well and robustly in real-world scenarios.
However, large parts of the local, regional, and export postharvest supply chain are decentralized, non-automated, and not digitalized from farm to retail. Therefore, stakeholders often do not have large annotated datasets readily available for their specific use case, defects, and cultivars that could be used to set up these models \citep{de-arteaga_machine_2018}. We argue that this domain has a particular need for computer vision models that work in the {\em low-data regime}, where images, annotations, or both are limited.

Recent computer vision research shows a trend toward {\em foundation models} that are pre-trained on large (often unlabeled) datasets, enabled by advances in self-supervised learning \citep{balestriero_cookbook_2023}.
From a practical view, these models promise to address low-data regimes as they are more easily adaptable to complex downstream tasks in a zero or few-shot manner \citep{knott_facilitated_2023}.
Most foundation models can be adapted to various tasks (e.g., classification, detection, segmentation, depth estimation) in a specific data modality (e.g., images, text, time series) or across modalities \citep[e.g.,][]{li_multimodal_2024, zhang_text2seg_2023}. Other foundation models are designed to excel in a certain task \citep{kirillov_segment_2023, depthanything} or domain problem \citep{israel_foundation_2023, shah_vint_2023, lu_visual-language_2024} within a modality.

In this work, we address the defect-based grading of fresh produce under constrained data and annotation availability, which occurs in many real-world scenarios, particularly in decentralized supply chains.
Our end goal is to count and estimate the size of visible surface defects (i.e., surface blemishes such as bruises and scars) on fresh produce, which is essential for grading. We formulate this problem as a panoptic segmentation task \cite{kirillov_panoptic_2019}.
While our experiments focus on bananas as a representative commodity, the proposed methodology is generalizable to other types of fresh produce with comparable grading criteria.

Our contribution is two-fold:
First, we address the issue of {\em label scarcity} that naturally occurs with dense computer vision tasks such as semantic or panoptic segmentation.
We will evaluate the {\em Segment Anything Model} (SAM) \cite{kirillov_segment_2023}, a promptable foundation model for image segmentation, for the weakly supervised generation of panoptic masks. More specifically, we evaluate how defect instance masks align with dense human annotations and look for systematic failure cases.
Second, using these predicted masks, we train a panoptic segmentation model on 476 images. We evaluate results against both human-annotated ground truth and masks generated by SAM.
This allows us to estimate how the error introduced by using SAM-generated masks propagates through the downstream task.
To our knowledge, this is one of the first works to combine foundation models like SAM with panoptic segmentation for defect quantification in the agricultural domain.

\section{Methods}

\begin{figure*}[t!]
    \centering
    \includegraphics[width=.96\textwidth]{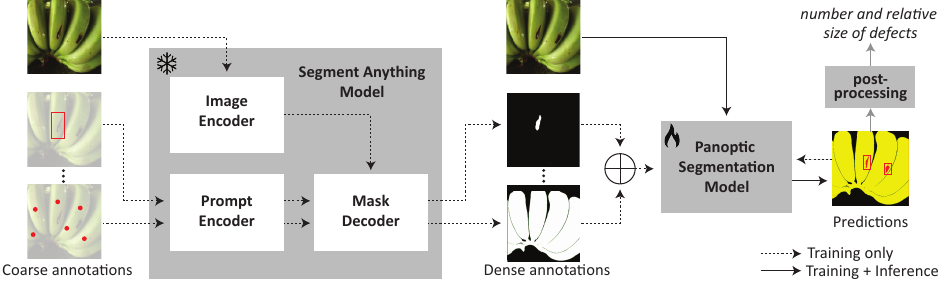}
    \caption{\textbf{Overview of our approach.} We utilize the Segment Anything Model (SAM), a promptable visual foundation model for image segmentation, to generate dense annotations (pixel-wise class and instance labels) from coarse annotations (bounding boxes and reference points) without any model training involved. These dense annotations otherwise require tedious hand-annotation. Using these newly generated labels, we train a panoptic segmentation model to identify surface defects on banana fruits, specifically bruises and scars. Additionally, the model can differentiate between foreground banana fruits and those in the background. This enables us to determine the number and size of visible surface defects from photographs of banana fruits. We validate our approach using a dataset of 476 images and 1440 annotated defects.}
    \label{fig:bananasam}
\end{figure*}

\subsection{Framing the problem}
\label{sec:framing}
Our work focuses on a common form of visual inspection relevant across a wide range of commodities: detecting surface defects, such as blemishes resulting from external impacts \citep{abbas_automated_2019}. Although grading criteria vary by commodity and market, we identify two key indicators that are consistently important: (a) the \emph{number} of visible blemishes and (b) the \emph{size} of each of those. Additionally, the \emph{type} of defect, while secondary, provides valuable insight into potential causes, aiding supply chain traceability and intervention.
We argue that an image segmentation approach is better suited for this task than regression or ordinal classification. Segmentation outputs are inherently interpretable and can be visually validated, which is crucial in quality assurance settings where transparency and accuracy are essential.

To estimate the number of defects, it is essential to distinguish between individual defect instances. Additionally, identifying banana fruits is necessary for relative size estimation (see \Cref{sec:num_and_size} for details); however, distinguishing between different banana fruit instances is not required beyond distinguishing foreground and background bunches in an image. Based on these requirements, we conclude that the \emph{panoptic segmentation} \cite{kirillov_panoptic_2019} framework—combining both semantic and instance segmentation—is well-suited to our task.
In principle, a standard semantic segmentation approach combined with postprocessing could also be used to separate instances (see \Cref{sec:postprocessing}), but panoptic segmentation offers a more integrated and efficient solution.

\subsection{Data collection}
\label{sec:dataset}

We collected 476 images of single banana bunches containing at least one visible bruise or scar in three different sorting and packing facilities in India, located in Tembhurni (Maharashtra state), Raver (Maharashtra state), and Kamrej (Gujarat state).
The images were taken with different smartphone cameras between harvesting and packing. 
To ensure independent data samples, each bunch was photographed only once.
The photos have realistic, non-standardized properties: the proximity to the fruits, lighting conditions, image sharpness, and resolution can vary. Also, there can be other bananas and objects occasionally in the background. Such conditions are typical for images taken in supply chains by inspectors, so we wanted our method to accommodate that.
We aim to first automatically detect, segment, and classify defects using a machine learning model. In the second step, we derive the number of visible surface defects and their relative size from these predictions.

\paragraph{Defect annotations} The images were annotated by a single expert using the open-source tool ``LabelStudio'' \citep{tkachenko_label_2020}. As we are interested in the number of defects, we annotated defects on an instance level, meaning that we annotated one bounding box per defect.
While a key property of our approach is to only use bounding boxes and generate instance masks via SAM, we also annotate defect masks for validation purposes.
Specifically, each defect was one of four categories: ``Old Bruise'', ``New Bruise'', ``Old Scar'', and ``New Scar''. 

Bruises typically result from blunt force or pressure, which can arise from various factors such as handling, packing, impacts, or weight.
Scars, on the other hand, are caused by sharp contact due to reasons like friction from leaves or stems, cuts from knives, crates, rough handling, laborers' nails, or the sharp tips and crowns of neighboring bunches.
Old defects are usually darker than new defects as they oxidize over time (see \Cref{fig:defect-examples}). 
In total, 1440 defects were annotated. (37 Old Bruise, 182 New Bruise, 387 Old Scar, 834 New Scar)
While the automated detection of generic defects without classification is already an impactful use case, we chose to add this categorization as it helps supply chain managers trace back the cause of the damage in the supply chain. 
It is important to note that there are cases that cannot be unambiguously assigned to one of those categories, even by expert annotators. However, a sub-goal of this work is to assess to which degree an automated classification is possible.
It is worth mentioning that there are plenty of other relevant surface defects reducing the quality of banana crops that are not included in this study, e.g., insect damage, speckles, sunburn, maturity stains, malformed fruits, broken necks, or chemical residues.

\paragraph{Banana annotations}
To obtain relative defect sizes, we also annotated segmentation masks of banana fruits. We decided to annotate bananas semantically only on a pixel level, omitting instance information for two reasons: First, as images often contain multiple bunches that are partly cropped or occluded, it is challenging to differentiate instances. Second, our use case does not require us to identify instances, as we are not interested in counting single bananas or bunches. We would rather distinguish focused foreground bunches from those that may appear in the background and, therefore, annotate two types of pixel classes: ``Foreground Banana'' and ``Background Banana''. All pixels not annotated as a banana of defect are assigned the ``Background'' class.
We annotated banana masks using LabelStudio and Segment Anything point prompts (see \Cref{sec:sam} for more details).

\subsection{Mask generation with SAM}
\label{sec:sam}

The Segment Anything Model (SAM) is a recently published foundation model for image segmentation \citep{kirillov_segment_2023}. It was trained on a large-scale dataset comprising diverse images from various domains, including natural scenes, urban environments, medical imagery, and other contexts.
It is the first model of its kind specifically tailored for generalized image segmentation and promises to facilitate the implementation of domain-specific segmentation models.

\Cref{fig:bananasam} (left) shows a high-level overview of SAM's components. The model consists of three major components: (1) a vision transformer \citep{dosovitskiy_image_2020} pre-trained with the Masked Autoencoder method \citep{he_masked_2021} that serves as an {\em image encoder}; (2) a prompt encoder that allows the user to specify bounding boxes or points to localize the region of interest in an image; (3) a mask decoder that generates segmentation masks based on the concatenated image and prompt embeddings.
This approach yields impressive results for zero-shot instance segmentation in various domains. However, compared to conventional segmentation models, it comes with two disadvantages that require further adaptations or post-processing to enable SAM for a typical image segmentation use case:
Firstly, SAM requires weak labels, such as bounding boxes or point coordinates, as input prompts.
Secondly, the output masks are generally uncategorized.

We explore the use of SAM in a zero-shot setting to generate dense panoptic masks from coarse annotations—specifically, bounding boxes for defects and point annotations for banana fruits—effectively serving as a semi-automated annotation tool. 
However, it is not immediately clear whether SAM-generated masks sufficiently align with human-annotated masks. Other than common objects in images, surface defects potentially have ambiguous boundaries that are subject to expert judgment.

\paragraph{Validating SAM masks}
We validate SAM masks in two ways: 
First, we directly evaluate the alignment between SAM-generated and human-annotated masks under different conditions, including ablations on the backbone model size and the resolution of the defect areas in the image (\Cref{sec:results_sam}).
Second, we assess the performance of our panoptic segmentation model against both human-annotated and SAM-generated masks (\Cref{sec:results_panoptic}). This allows us to draw conclusions about the extent of error propagation resulting from inaccurate mask generation.
More specifically, we report panoptic evaluation using three different combinations of mask sources for training and validation data:
(a) both training and validation use hand-annotated defect masks; 
(b) both training and validation use SAM-generated defect masks; and 
(c) training uses SAM-generated masks while validation uses hand-annotated masks. By comparing the evaluation metrics across these three variations, we can quantify the trade-off between the time-saving benefits of automated mask generation and the quality of the resulting output.


\subsection{Image segmentation}
\label{sec:image-segmentation}

In \Cref{sec:framing}, we argued why we frame this use case as a hand as a panoptic segmentation problem.
This approach yields a dense semantic segmentation mask for the whole image while instance identification is provided for certain specified categories only \citep{cheng_per-pixel_2021, cheng_masked-attention_2022, kirillov_panoptic_2019-1, kirillov_panoptic_2019}.
In our study, we declare defects as ``things'' (countable) and bananas as ``stuff'' (non-countable) as we are not interested in distinguishing single banana fruits.
We train a Maskformer~\citep{cheng_per-pixel_2021} pre-trained on the ADE20k dataset \citep{zhou_scene_2017} in a standard supervised learning workflow (\Cref{fig:bananasam}, right).

\paragraph{Evaluation metrics}
We evaluate the final model using metrics from all three paradigms: semantic, instance, and panoptic segmentation.
Semantic segmentation maps are evaluated against a groundtruth using the Intersection-over-Union (IoU) score for a single class or the mean IoU (mIoU) score for all classes in a dataset \citep{everingham_pascal_2012, lin_microsoft_2014}.
For instance detections, we follow the established practice to report average precision at different $IoU$ threshold ($AP@50$, $AP@75$) \citep{everingham_pascal_2012}. $AP$ is equivalent to $AP^{IoU=.50:.05:.95}$, averaged over ten equidistant $IoU$ thresholds between $0.50$ and $0.95$ as proposed by \citet{lin_microsoft_2014}.
For the average recall metric ($AR$), we follow the definition of \citet{hosang_what_2015} evaluating a range of IOU thresholds.
In addition, we use {\em Panoptic Quality} ($PQ$) as introduced by \citet{kirillov_panoptic_2019} as an integrated metric for panoptic segmentation. 

\paragraph{Cross-validation}
Given the relatively small size of our training dataset, it is important to ensure that our reported results are robust. To achieve this, we use five-fold cross-validation instead of a static validation split. This approach involves dividing the dataset into five equally sized parts. We repeat each experiment five times, with one partition serving as the validation set while the remaining four partitions are used to train the model in each iteration. The evaluation metrics are then reported as the mean and standard deviation across these five experiments.

Further details on model training, such as hyperparameters and data augmentation, can be found in \Cref{appx:implementation-details}.

\subsection{Postprocessing}
\label{sec:postprocessing}
In some cases, it can be ambiguous whether to treat closely located defects as a single defect or multiple defects. To address this and ensure more consistent defect separation, we introduce a simple algorithm-based postprocessing method with the following three steps: 
\begin{enumerate}
    \item Combine all defect instances of the same type into a single binary map (skip if a semantic segmentation model is used).
    \item Use a connected components algorithm \citep{opencv_open_2015} to generate new instances where no pixels are connected.
    \item Apply a dilation operation to expand all instance masks by $d$ pixels (we choose $d=5$) in all directions. This expanded auxiliary mask is then used to identify overlapping instances, which are subsequently merged. In other words, defect instances of the same class within a distance of $2d$ pixels are merged.
\end{enumerate}
Note that after step~1, we have regular semantic segmentation representation, and thus, using steps 2 and 3, one could also obtain panoptic results based on standard semantic segmentation outputs. However, during initial experimentation, we found that training a Maskformer~\citep{cheng_per-pixel_2021} yields better results than a comparable semantic segmentation model (Segformer~\cite{xie_segformer_2021}). Also, we consider this postprocessing as optional when a panoptic model is used.

\subsection{Deriving number and size of defects}
\label{sec:num_and_size}

For many real-world grading applications, one is interested in the {\em number} and {\em size} of defects, which can both be derived (or approximated) from panoptic masks. While the first can be estimated by counting the number of defect instances, an accurate metric size estimation from a single image is generally impossible due to unknown scale, perspective distortion, and lack of depth information. Thus, we use the {\em relative} defect size, simplified as the number of visible defect pixels divided by the number of visible foreground banana pixels, as a proxy.

\section{Results}
\label{sec:results}

\subsection{SAM defect masks} 
\label{sec:results_sam}

\begin{figure}[t!]
    \centering
    \includegraphics[width=\linewidth]{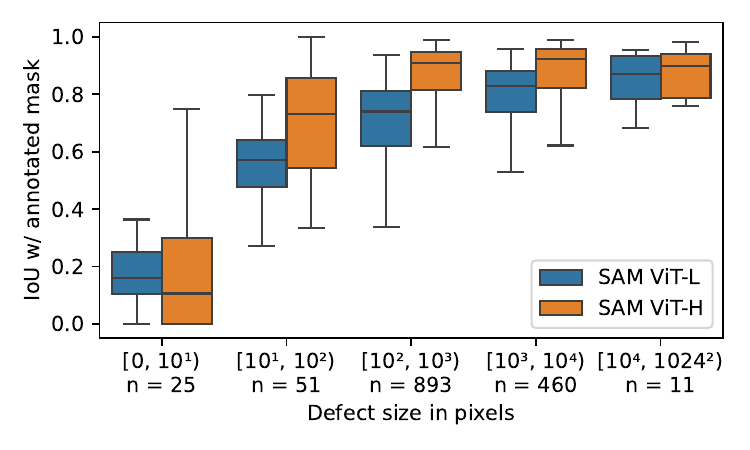}
    \caption{\textbf{Agreement between human-annotated and SAM-predicted masks by defect size (number of pixels).} The x-axis shows binned size categories for defect sizes in pixels (per annotated mask), while $n$ denotes the number of samples in each bin. The y-axis shows the agreement between annotated and predicted masks (IoU). SAM fails to align with human annotations for small ($<\!100$ pixels) and very small ($<\!10$ pixels) defects. To understand the failure cases of larger defects ($>\!100$ pixels), we added exemplary visualizations in \Cref{fig:sam-failure}. Those cases often show thin, long scars.
    We can also see that a larger SAM model size (ViT-H) leads to consistently higher agreement with human annotations across all defect sizes. An additional analysis of the impact of SAM model sizes can be found in \Cref{fig:sam-ious}.
    }
    \label{fig:sam-size}
\vspace{5mm}
    \includegraphics[width=\linewidth]{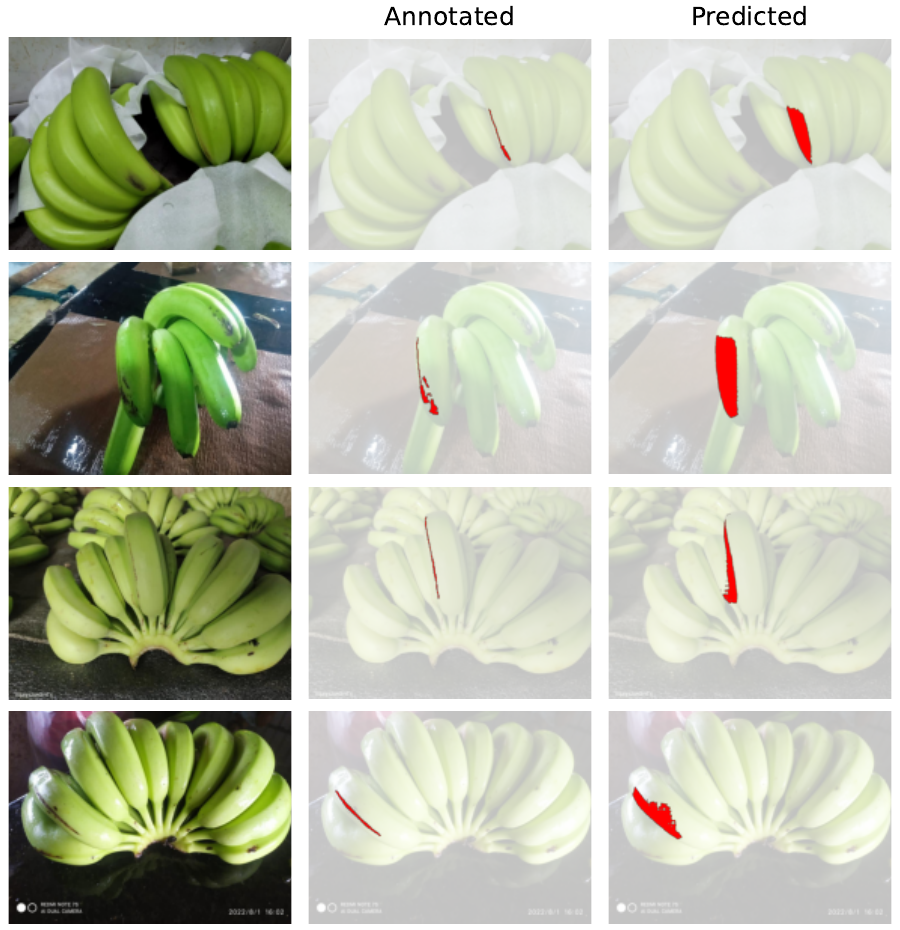}
    \caption{\textbf{SAM failure cases.} Exemplary examples where SAM (ViT-L) fails to align with the annotated masks. We picked examples with the lowest IoU where the annotated mask size is $>\!100$ pixels. The examples shown illustrate that long, thin scars are difficult to capture. Only one defect per image is shown.}
    \label{fig:sam-failure}
\end{figure}

How well do SAM-predicted masks align with human annotations?
We use SAM in a zero-shot manner without any fine-tuning. Thus, it is not self-evident that all defects with different shapes and sizes align closely with human annotations. Also, the choice of the SAM backbone might have an influence on predictions as model capacity usually correlates with model accuracy.
\Cref{fig:sam-size} compares the agreement between hand-annotated and SAM-generated masks for two different SAM backbone sizes (ViT-L and ViT-H) and for different defect sizes defined by the number of pixels in the annotated mask.
First, we can see that a larger SAM model size (ViT-H) leads to consistently higher agreement with human annotations across all defect sizes. This finding is confirmed by an additional analysis where we plot the IoU distributions for three different SAM backbones (\Cref{fig:sam-ious}). However, we see that the alignment is generally high, and failure cases are long-tail outliers for all model backbones.
A second observation from \Cref{fig:sam-size} is that SAM fails to align with human annotations for small ($<\!100$ pixels) and very small ($<\!10$ pixels) defects. To understand the failure cases of larger defects ($>\!100$ pixels), we added exemplary visualizations in \Cref{fig:sam-failure}. In those cases, which often show long, thin scars, SAM produces masks that are too large.

\subsection{Panoptic predictions}
\label{sec:results_panoptic}

\begin{table*}[t]
\centering
\caption{\textbf{Image segmentation metrics.} All results are reported via their mean and standard deviation over the five different cross-validation splits. We compare different setups where target defect masks are either annotated (``Anno.'') or generated by SAM (-L/-H refers to the size of the ViT backbone). Expectedly, when Maskformer is trained on SAM-generated masks, the defect IoU is slightly lower when evaluating annotated masks. In the same comparison, recall and precision are degrading less strongly. Postprocessing (``PP'') significantly improves precision and recall of defects. Using SAM-generated instead of annotated training masks, leads to identical PQ ($\simsym0.779$), while mIoU is slightly worse ($\simsym0.808$ vs $\simsym0.810$). A larger SAM backbone size does not lead to an improvement in downstream segmentation quality. The configuration highlighted in \colorbox{LightBlue}{blue} is used for subsequent visualizations.}
\begin{adjustbox}{width=\textwidth}
\label{tab:results-single}
\begin{tabular}{lccc|ccccc|cc|cc}
\toprule
\multicolumn{2}{c}{} & \multicolumn{2}{c}{Defect masks} & \multicolumn{5}{c}{Defects} & \multicolumn{2}{c}{Fruits} & \multicolumn{2}{c}{Overall} \\
\cmidrule(lr){3-4} \cmidrule(lr){5-9} \cmidrule(lr){10-11} \cmidrule{12-13} 
Model & PP & train & val & \multicolumn{1}{c}{AP} & \multicolumn{1}{c}{AP@50} & \multicolumn{1}{c}{AP@75} & \multicolumn{1}{c}{AR} & \multicolumn{1}{c}{IoU} & \multicolumn{1}{c}{IoU FG} & \multicolumn{1}{c}{IoU BG} & \multicolumn{1}{c}{mIoU} & \multicolumn{1}{c}{PQ} \\
 \midrule 
Maskformer &  & Anno. & Anno. & $.153 \pm .040$ & $.263 \pm .066$ & $.163 \pm .050$ & $.262 \pm .051$ & $.563 \pm .037$ & $.956 \pm .006$ & $.750 \pm .050$ & $.810 \pm .018$ & $.764 \pm .010$ \\
Maskformer & \checkmark & Anno. & Anno. & $.223 \pm .030$ & $.413 \pm .059$ & $.222 \pm .040$ & $.397 \pm .034$ & $.563 \pm .037$ & $.956 \pm .006$ & $.750 \pm .050$ & $.810 \pm .018$ & $.779 \pm .010$ \\
\\
Maskformer &  & SAM-L & Anno. & $.129 \pm .011$ & $.231 \pm .019$ & $.134 \pm .017$ & $.243 \pm .024$ & $.559 \pm .024$ & $.952 \pm .007$ & $.746 \pm .056$ & $.808 \pm .018$ & $.760 \pm .008$ \\
\rowcolor{LightBlue} Maskformer & \checkmark & SAM-L & Anno. & $.217 \pm .020$ & $.405 \pm .038$ & $.211 \pm .024$ & $.393 \pm .035$ & $.562 \pm .028$ & $.952 \pm .007$ & $.746 \pm .056$ & $.808 \pm .018$ & $.779 \pm .008$ \\
Maskformer &  & SAM-L & SAM-L & $.137 \pm .015$ & $.232 \pm .015$ & $.147 \pm .023$ & $.253 \pm .025$ & $.574 \pm .022$ & $.952 \pm .007$ & $.746 \pm .056$ & $.811 \pm .018$ & $.762 \pm .006$ \\
Maskformer & \checkmark & SAM-L & SAM-L & $.231 \pm .023$ & $.409 \pm .050$ & $.235 \pm .026$ & $.410 \pm .039$ & $.576 \pm .025$ & $.952 \pm .007$ & $.746 \pm .056$ & $.812 \pm .018$ & $.782 \pm .007$ \\
\\
Maskformer &  & SAM-H & Anno. & $.150 \pm .016$ & $.257 \pm .027$ & $.160 \pm .021$ & $.251 \pm .033$ & $.547 \pm .063$ & $.952 \pm .008$ & $.733 \pm .045$ & $.801 \pm .023$ & $.759 \pm .007$ \\
Maskformer & \checkmark & SAM-H & Anno. & $.222 \pm .028$ & $.418 \pm .056$ & $.211 \pm .031$ & $.384 \pm .048$ & $.548 \pm .064$ & $.952 \pm .008$ & $.733 \pm .045$ & $.801 \pm .023$ & $.777 \pm .010$ \\
Maskformer &  & SAM-H & SAM-H & $.158 \pm .022$ & $.255 \pm .024$ & $.175 \pm .031$ & $.258 \pm .033$ & $.573 \pm .067$ & $.952 \pm .008$ & $.733 \pm .045$ & $.807 \pm .024$ & $.763 \pm .007$ \\
Maskformer & \checkmark & SAM-H & SAM-H & $.242 \pm .031$ & $.422 \pm .048$ & $.251 \pm .043$ & $.403 \pm .058$ & $.574 \pm .068$ & $.952 \pm .008$ & $.733 \pm .045$ & $.808 \pm .024$ & $.783 \pm .011$ \\

\bottomrule
\end{tabular}
\end{adjustbox}
\end{table*}

Next, we assess image segmentation results on validation images using the metrics outlined in \Cref{sec:image-segmentation}.
\Cref{tab:results-single} presents these metrics across various experimental setups:
In the first experiment, we used hand-annotated defect masks for both training and validation, achieving an average panoptic quality (PQ) of 77.9\% after applying postprocessing. This serves as our baseline for assessing how segmentation quality changes when replacing hand-annotated masks with SAM-generated ones.
In the second experiment, we employ SAM-generated masks for both training and validation. The results show a slightly higher PQ score of 78.2\% (ViT-L backbone).
In the final experiment, SAM-generated masks are used for training, while hand-annotated masks are used for validation. 
One can see a slight drop in detection metrics (AP: $-0.6\%$, AR: $-0.4\%$, IoU: $-0.1\%$), while the overall PQ score is almost identical when SAM-generated instead of hand-annotated masks are used for training.

While we previously showed that the larger SAM variant (ViT-H) has a higher defect mask agreement with hand annotations (see \Cref{fig:sam-size}), this does {\em not}, somewhat counterintuitively, lead to better segmentation metrics. This suggests that the downstream model (MaskFormer, in our case) can compensate for a certain degree of inaccuracy in the training masks.

\begin{figure*}[htp]
    \centering
    \includegraphics[width=.9\linewidth]{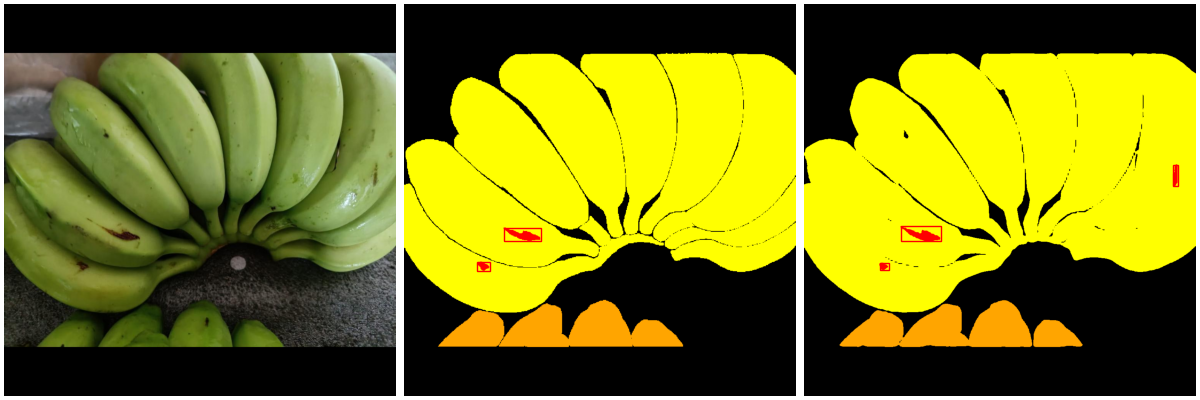}
    \includegraphics[width=.9\linewidth]{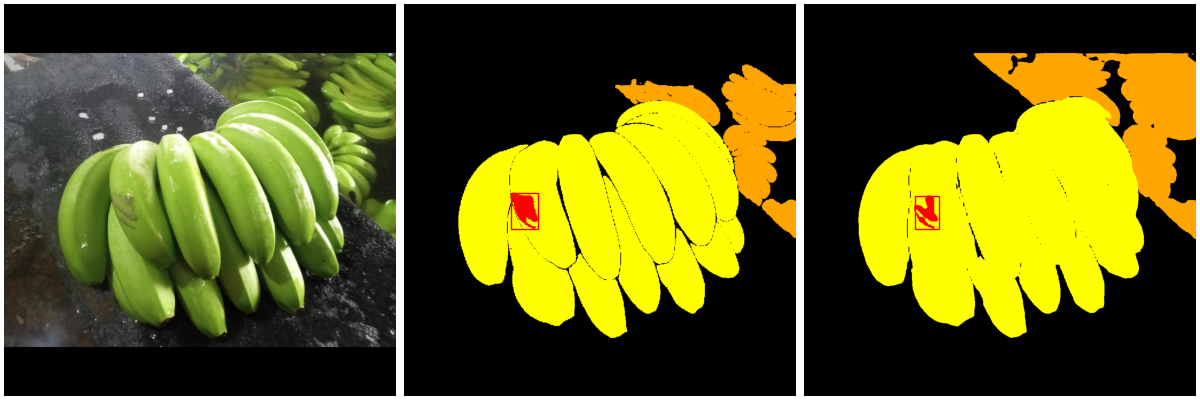}
    \includegraphics[width=.9\linewidth]{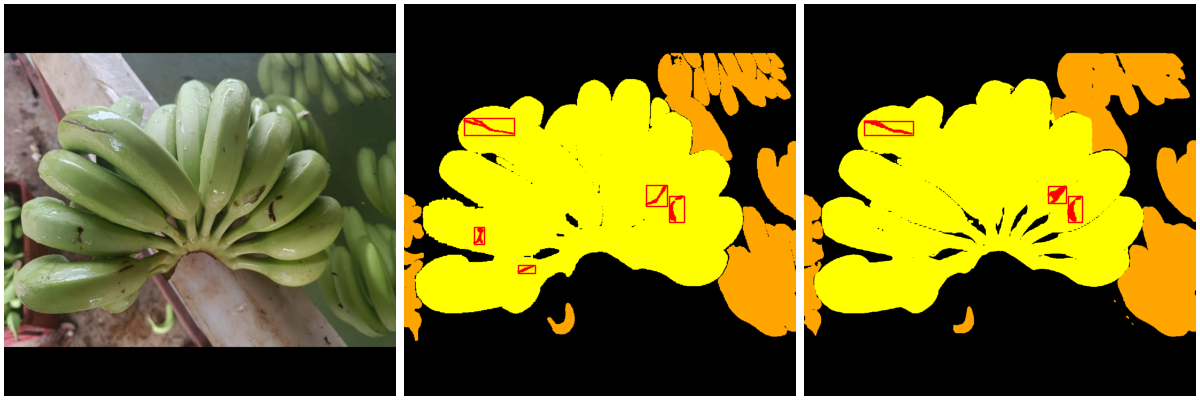}
    \includegraphics[width=.9\linewidth]{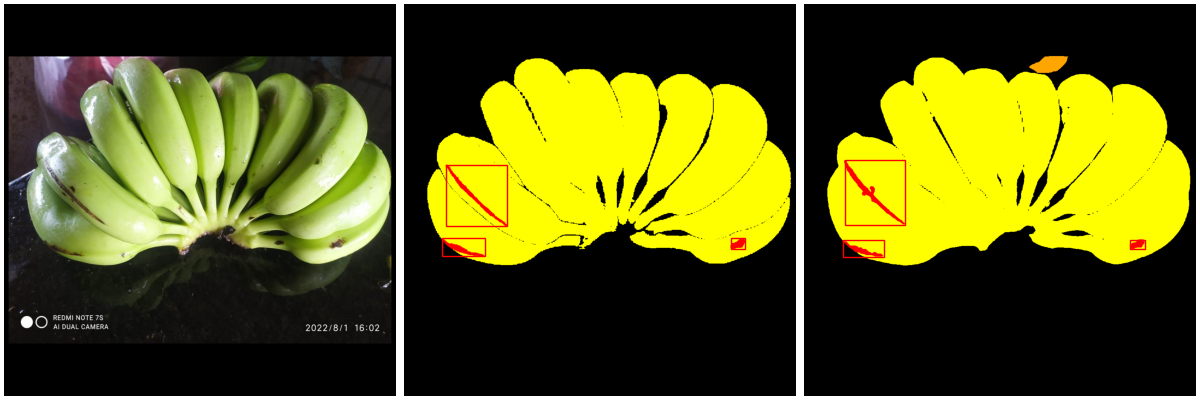}
    \caption{\textbf{Example visualizations of annotated vs predicted masks.} Left: Input Image, Mid: Annotation, Right: Maskformer Prediction. Red rectangles enclose defect instances.
    Segments are color-coded as follows: \colorbox{Yellow}{Foreground Banana}, \colorbox{Orange}{Background Banana}, \colorbox{Red}{Defect},
    }
    \label{fig:examples}
\end{figure*}

Across all settings, IoU scores for segmenting foreground bananas are consistently high ($>95\%$), while those for background bananas are noticeably lower ($\simsym74\%$). This suggests that the model rarely confuses the two classes—if it did, both scores would be low. Instead, we conclude that background bananas are occasionally missed in the annotations. Qualitative comparisons support this interpretation (see, for example, \Cref{fig:examples}, second and fourth row).

Furthermore, the postprocessing described in \Cref{sec:postprocessing} yields a significant improvement in AP across all settings.

The panoptic results can be qualitatively analyzed through the example visualizations in \Cref{fig:examples} showing the segmentation using a single generic defect class. In general, the masks appear sharp, particularly for bananas, and the model reliably distinguishes between fruits in the foreground or center of the image and those in the background. While the detected defects do not always align with the annotations, they are generally plausible. The model often identifies blemishes that were not annotated, likely because they were deemed too minor to be significant. This underscores the challenge of defining ground-truth defect annotations in these images.

\begin{figure}[t]
    \centering
    \includegraphics[width=.75\linewidth]{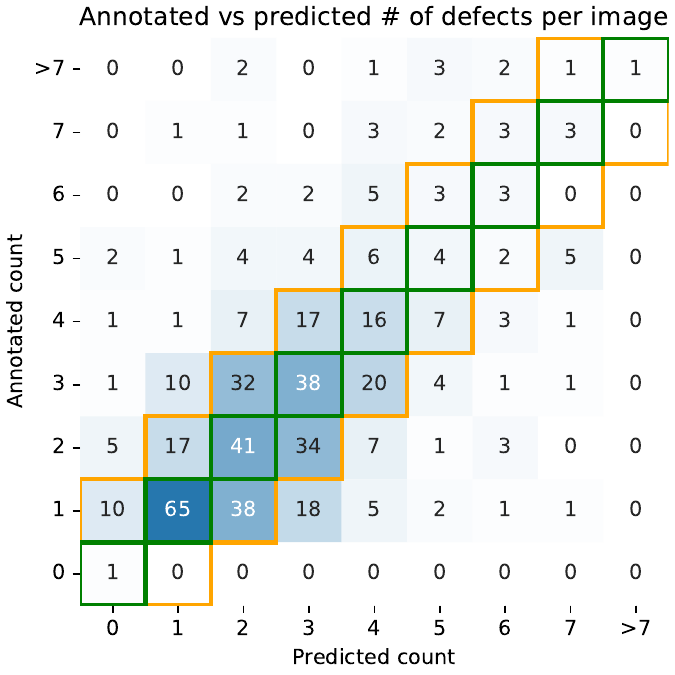}
    \caption{\textbf{Annotated versus predicted defect counts per image.} We predict the exact number of defects 36.2\% of the time (green squares). 76.2\% of the time, we predict correctly within a $\pm1$ tolerance (orange squares). Generally, our model tends to predict more defects in the images compared to the annotations.}
    \label{fig:n-defects-matrix}
\end{figure}

\begin{figure}[t]
    \centering
    \includegraphics[width=.75\linewidth]{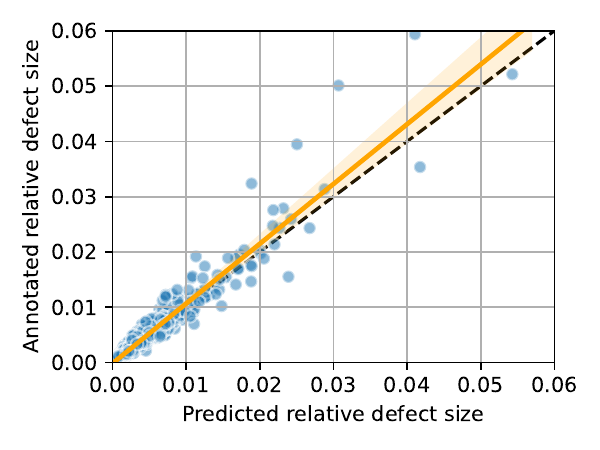}
    \caption{\textbf{Annotated versus predicted defect sizes.} We pairwise match those defect instances with the highest IoU agreement between annotation and estimation (minimum 0.5) and calculate their sizes relative to the corresponding foreground banana masks (non-matchable defects are excluded in this analysis). Relative size prediction is generally accurate with a Pearson correlation of $r=0.96, n=793$. Each blue dot is one pair of defects, the orange line is a fitted regression line, and the black dashed line is the $x=y$ diagonal.}
    \label{fig:size-defects}
\end{figure}

Lastly, we also conducted experiments using four distinct defect types instead of a single defect class. We found that defect categorization is very unreliable and attribute this mainly to ambiguous annotations. Thus, we do not recommend using our dataset for defect categorization at this point (see \Cref{appx:multidefect} for a more detailed discussion). 

\subsection{Number and size of defects}

Panoptic segmentation masks can simply be used to derive the number and size of defects. We compare the number of defects detected per image (\Cref{fig:n-defects-matrix}) and the relative defect sizes (\Cref{fig:size-defects}) by our best model (x-axes) and the human annotations (y-axes). 
The model matches the exact number of defects in 36.2\% of cases and is off by one in 76.2\% of cases. 
For relative size analysis, we pairwise match those defect instances with the highest IoU agreement between annotation and estimation. If no agreement with at least 0.5 IoU can be established, we consider predicted defects non-matchable and exclude them from this analysis. Relative sizes are calculated by dividing the number of pixels of a specific defect instance by the aggregated number of pixels assigned to either the foreground banana class or any defect.
We generally see a high agreement in predicted vs. annotated relative sizes ($r=0.96)$, while the model tends to slightly underestimate sizes, visible by the different slopes of the regression line (orange) and $x=y$ (black dashed) lines in \Cref{fig:size-defects}. 

\section{Discussion and Limitations}

In this work, we lay the foundation for the automated, defect-based grading of fresh produce using limited samples and annotations.  
We present a carefully curated dataset comprising 476 real-world images of banana bunches and 1,440 annotated surface defects covering a wide range of sizes, shapes, and categories.

Further, we highlight the trade-off when using SAM for generating dense image masks from coarse annotations. Specifically, we estimate that the manual annotation workload can be reduced by at least a factor of 10 while image segmentation metrics for the downstream panoptic segmentation task experience only a moderate decrease (Defect IoU 56.3\%\textrightarrow56.2\%), Defect AP 22.3\%\textrightarrow21.7\%, Defect AR 39.7\%\textrightarrow39.3\%).
We train a panoptic segmentation model and propose both a postprocessing algorithm and a procedure to extract grading-relevant metrics—such as defect count and relative size—from the model’s predictions. 
While it can be questioned if the current model is sufficient for practical deployment, it shows promise. Compared to human annotations, the model predicts the exact number of defects in 36.2\% of cases and is off by at most one defect in 76.2\% of cases. Although part of this discrepancy can be attributed to model limitations, we identify several avenues for improvement for future work, which we outline in \Cref{sec:future-work}.

Lastly, it is important to note that countable and sizable defects represent only a subset of relevant grading criteria. For bananas, other defect types—such as speckles, sunburn, chemical residues, or malformed shapes—may require fundamentally different approaches, such as classification or ordinal regression.

\section{Future Work}
\label{sec:future-work}

\paragraph{Improving model performance}
We see several ways to improve model performance on our dataset:
\begin{itemize}
    \item \textit{Model choice:} The landscape of deep learning model architectures evolves rapidly. Our approach will likely benefit from better SAM variants and improved panoptic segmentation architectures. SAM~2 \cite{ravi_sam_2024} and Mask2Former \cite{cheng_masked-attention_2022} are the logical next candidates to be evaluated in our framework.
    \item \textit{Fine-tune SAM:} Instead of using SAM in a zero-shot manner, one could fine-tune the model on domain data to achieve better segmentation masks for domain-specific instances. This was already successfully done in different contexts \cite[see e.g.,][]{israel_foundation_2023}. We hypothesize that this will help with the issues of thin, long scars that can be seen in \Cref{fig:sam-failure}.
    \item \textit{Categorized segmentation with SAM:} Instead of our current two-model setup, adapting SAM to directly support categorized semantic or panoptic segmentation in a zero- or few-shot manner would provide a more elegant solution. Future research will likely explore such extensions.
    \item As we have highlighted in \Cref{fig:sam-size}, SAM fails with very small defect sizes. We expect that small defect predictions will improve when increasing the used image resolution (we use 1024\textsuperscript{2}).
\end{itemize}

\paragraph{Improving data set creation}
We also want to highlight the issue of ambiguity in expert annotations for defect segmentation and, in particular, categorization. We would like to encourage a broader discussion on this challenge. One potential improvement would be to involve multiple annotators and quantify their agreement. The value of using (noisy) annotations from multiple annotators, especially for non-trivial labeling tasks, has been well documented in the machine learning literature \citep{welinder_multidimensional_2010, sheng_get_2008}. We, therefore, recommend that future work adopt more sophisticated annotation protocols involving multiple experts, using inter-annotator disagreement as a baseline for evaluating machine learning models.

\section*{Acknowledgments}
This work was partially funded by the project ``Scaling up Your Virtual Cold Chain Assistant'', commissioned by the German Federal Ministry of Economic Cooperation and Development (BMZ) on behalf of the German International Cooperation Agency (GIZ).

\FloatBarrier

\bibliographystyle{ieeenat_fullname}
\bibliography{references} 

\clearpage
\appendix
\onecolumn
\setcounter{figure}{0}    
\setcounter{table}{0} 
\renewcommand\thefigure{S.\arabic{figure}} 
\renewcommand\thetable{S.\arabic{table}}
\appendix 
\begin{center}
    \begin{minipage}{\textwidth}
        \centering
        \fontsize{14}{18}\selectfont
        \textbf{Supplementary Materials}
    \end{minipage}
\end{center}
\vspace{6pt}

\section{Dataset visualizations}

\begin{figure}[h]
    \centering
    \includegraphics[width=\linewidth]{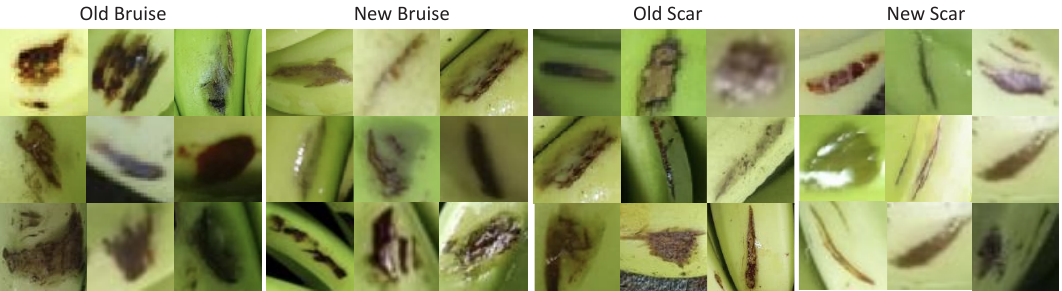}
    \caption{\textbf{Defect examples.} Example images of banana surface defects we aim to detect in this study. Bruises usually result from a dull impact, while scars are caused by a sharp impact. Old defects are usually darker than new defects as they oxidize over time.}
    \label{fig:defect-examples}
\end{figure}

\FloatBarrier
\section{Implementation details}
\label{appx:implementation-details}

All images were resized and padded to $1024^2$ pixels resolution. Training images were augmented with a 50\% Random Horizontal Flip and with additional random color-based augmentations (see \Cref{tab:color-augment}) to increase model robustness.

\Cref{tab:hyperparameters} shows the hyperparameters used to train the Maskformer models. We evaluated the model every five epochs on the validation set and saved the model with the highest Panoptic Quality. All models were trained on a single NVIDIA RTX A5000 GPU with 24GB of memory.

\begin{table}[h]
    \caption{\textbf{Random color augmentations applied to training images.} Values are uniformly sampled from the specified ranges.}
    \label{tab:color-augment}
    \centering
    \begin{tabular}{ll}
    \toprule
    Property & Sampling range \\
    \midrule
     brightness    & [0.9, 1.1]  \\
     contrast    & [0.9, 1.1]  \\
     saturation    & [0.9, 1.1] \\
     hue    & [-0.05, 0.05] \\
     \bottomrule
    \end{tabular}
\end{table}

\begin{table}[h]
    \centering
    \caption{\textbf{Hyperparameters for Maskformer training.}}
    \begin{tabular}{ll}
       \toprule
        Batch size                      & 2                             \\
        Epochs                          & 100                           \\
        Evaluation frequency            & every 5 epochs                \\
        Optimizer                       & Adam~\citep{kingma_adam_2014}   \\
        Learning rate                   & $5 \times 10^{-5}$            \\
        Learning rate schedule          & constant                      \\
        \bottomrule
    \end{tabular}
    \label{tab:hyperparameters}
\end{table}

\FloatBarrier
\clearpage
\section{Additional SAM evaluations}

\begin{figure}[ht]
    \centering
    \includegraphics[width=.4\linewidth]{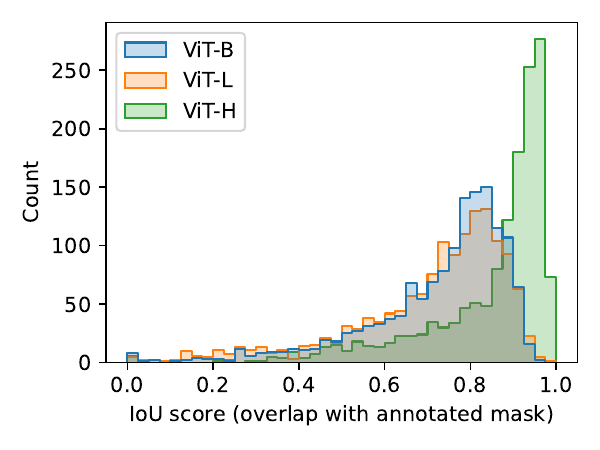}
    \caption{\textbf{Distribution of IoU values when comparing annotated and SAM-generated defect masks.} While the ViT-B and ViT-L variants show a similar distribution, using the largest model size (ViT-H) leads to significantly higher overlap with hand-annotated masks.}
    \label{fig:sam-ious}
\end{figure}

\FloatBarrier
\section{Results using four defect categories}
\label{appx:multidefect}

In this section, we describe our results when using four different defect classes instead of a single joint category. Our results clearly show that while the detection and segmentation of bruises and scars work well, a reliable categorization into one of the four predefined classes is not possible with the current approach and dataset.
We hypothesize that this limitation stems from one or more of the following factors:
\begin{enumerate}
    \item Ambiguous annotation: The classification of defects across the four predefined categories may be subjective to a large degree. The distinction between ``old'' and ''new`` defects is not always clear-cut, as the transition is gradual. Additionally, distinguishing bruises from scars can be ambiguous, especially for non-experts (see \Cref{fig:defect-examples}).    
    \item Image resolution: We use $1024^2$ pixel resolution for images. While this is generally considered high for machine learning tasks, defects can be small and, thus, only be represented by a few pixels, making them harder to categorize.
    
    \item The four defect types are unevenly represented in our dataset (37/182/387/834). A more balanced distribution of categories and a larger number of defect samples are likely to improve categorization accuracy. We recommend collecting more examples from the two underrepresented classes for future work.
\end{enumerate}

\begin{table*}[ht]
\centering
\caption{\textbf{Results using multiple defect categories.} It is evident that our models are unable to categorize defect types. Most likely due to ambiguous annotations and/or limited training data. The configuration highlighted in \colorbox{LightBlue}{blue} is used for the visualizations in \Cref{fig:examples-multi}.}
\label{tab:results-multi}
\begin{adjustbox}{width=\textwidth}
\begin{tabular}{lccc|cccccccc|ccc}
\toprule
\multicolumn{4}{c}{} & \multicolumn{8}{c}{Defects} & \multicolumn{3}{c}{} \\
\cmidrule(lr){3-12} 
\multicolumn{2}{c}{} & \multicolumn{2}{c}{Defect masks} & \multicolumn{2}{c}{Old Bruise} & \multicolumn{2}{c}{New Bruise} & \multicolumn{2}{c}{Old Scar} & \multicolumn{2}{c}{New Scar} & \multicolumn{3}{c}{Overall} \\
\cmidrule(lr){3-4} \cmidrule(lr){5-6} \cmidrule(lr){7-8} \cmidrule(lr){9-10} \cmidrule(lr){11-12} \cmidrule(lr){13-15} 
Model & PP & train & val & \multicolumn{1}{c}{AP} & \multicolumn{1}{c}{IoU} & \multicolumn{1}{c}{AP} & \multicolumn{1}{c}{IoU} &  \multicolumn{1}{c}{AP} & \multicolumn{1}{c}{IoU} &  \multicolumn{1}{c}{AP} & \multicolumn{1}{c}{IoU} & \multicolumn{1}{c}{mAP} & \multicolumn{1}{c}{mIoU} & \multicolumn{1}{c}{PQ} \\
 \midrule
Maskformer &  & Anno. & Anno. & $.031 \pm .059$ & $.028 \pm .043$ & $.043 \pm .024$ & $.157 \pm .037$ & $.036 \pm .018$ & $.225 \pm .097$ & $.050 \pm .018$ & $.316 \pm .054$ & $.040 \pm .019$ & $.482 \pm .024$ & $.471 \pm .023$ \\
Maskformer & \checkmark & Anno. & Anno. & $.030 \pm .059$ & $.013 \pm .018$ & $.088 \pm .039$ & $.114 \pm .035$ & $.034 \pm .016$ & $.223 \pm .083$ & $.017 \pm .009$ & $.066 \pm .021$ & $.042 \pm .022$ & $.436 \pm .018$ & $.431 \pm .015$ \\
\\
\rowcolor{LightBlue} Maskformer &  & SAM-L & Anno. & $.039 \pm .055$ & $.060 \pm .080$ & $.034 \pm .016$ & $.175 \pm .074$ & $.045 \pm .034$ & $.179 \pm .022$ & $.066 \pm .018$ & $.345 \pm .046$ & $.046 \pm .012$ & $.487 \pm .018$ & $.477 \pm .013$ \\
Maskformer & \checkmark & SAM-L & Anno. & $.032 \pm .049$ & $.031 \pm .038$ & $.074 \pm .017$ & $.124 \pm .051$ & $.038 \pm .021$ & $.161 \pm .009$ & $.032 \pm .014$ & $.104 \pm .039$ & $.044 \pm .015$ & $.437 \pm .018$ & $.435 \pm .019$ \\
Maskformer &  & SAM-L & SAM-L & $.037 \pm .053$ & $.063 \pm .081$ & $.036 \pm .016$ & $.178 \pm .075$ & $.045 \pm .033$ & $.182 \pm .023$ & $.071 \pm .019$ & $.353 \pm .049$ & $.047 \pm .012$ & $.489 \pm .018$ & $.478 \pm .013$ \\
Maskformer & \checkmark & SAM-L & SAM-L & $.032 \pm .049$ & $.027 \pm .034$ & $.080 \pm .018$ & $.127 \pm .051$ & $.036 \pm .020$ & $.163 \pm .010$ & $.034 \pm .015$ & $.108 \pm .041$ & $.046 \pm .015$ & $.437 \pm .018$ & $.436 \pm .019$ \\
\\
Maskformer &  & SAM-H & Anno. & $.019 \pm .026$ & $.074 \pm .078$ & $.043 \pm .020$ & $.157 \pm .092$ & $.031 \pm .021$ & $.252 \pm .063$ & $.056 \pm .016$ & $.318 \pm .047$ & $.037 \pm .014$ & $.495 \pm .010$ & $.474 \pm .010$ \\
Maskformer & \checkmark & SAM-H & Anno. & $.032 \pm .056$ & $.030 \pm .029$ & $.093 \pm .022$ & $.145 \pm .074$ & $.025 \pm .029$ & $.230 \pm .041$ & $.029 \pm .011$ & $.100 \pm .050$ & $.045 \pm .016$ & $.453 \pm .013$ & $.449 \pm .009$ \\
Maskformer &  & SAM-H & SAM-H & $.025 \pm .035$ & $.092 \pm .105$ & $.049 \pm .022$ & $.163 \pm .091$ & $.037 \pm .024$ & $.264 \pm .064$ & $.062 \pm .018$ & $.328 \pm .047$ & $.043 \pm .015$ & $.501 \pm .010$ & $.482 \pm .011$ \\
Maskformer & \checkmark & SAM-H & SAM-H & $.045 \pm .078$ & $.029 \pm .027$ & $.106 \pm .022$ & $.149 \pm .073$ & $.027 \pm .028$ & $.239 \pm .041$ & $.034 \pm .010$ & $.104 \pm .051$ & $.053 \pm .019$ & $.455 \pm .013$ & $.456 \pm .010$ \\

\bottomrule
\end{tabular}
\end{adjustbox}
\end{table*}

\begin{figure*}[ht]
    \centering
    \includegraphics[width=.9\linewidth]{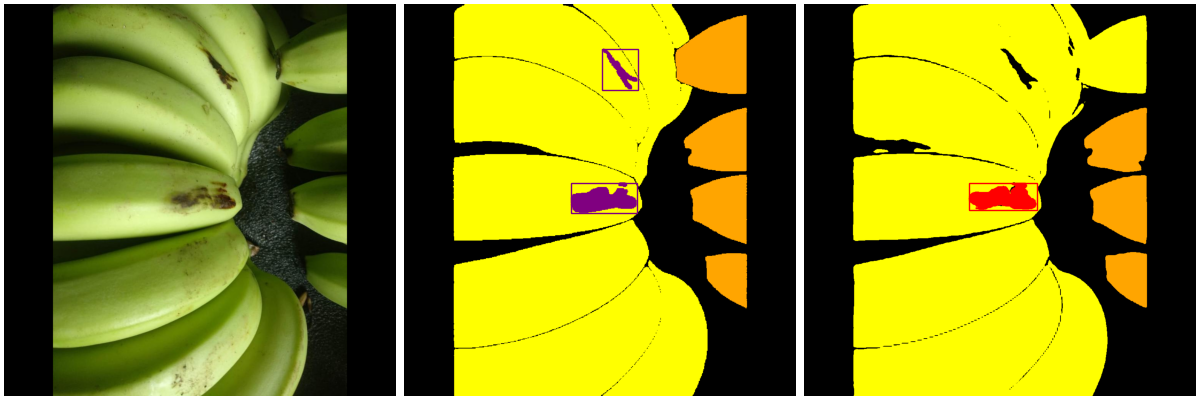}
    \includegraphics[width=.9\linewidth]{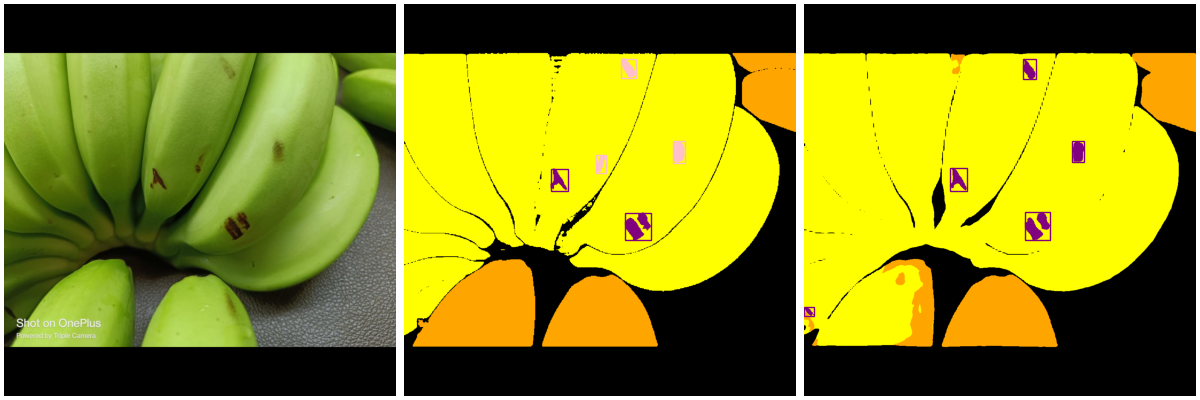}
    \includegraphics[width=.9\linewidth]{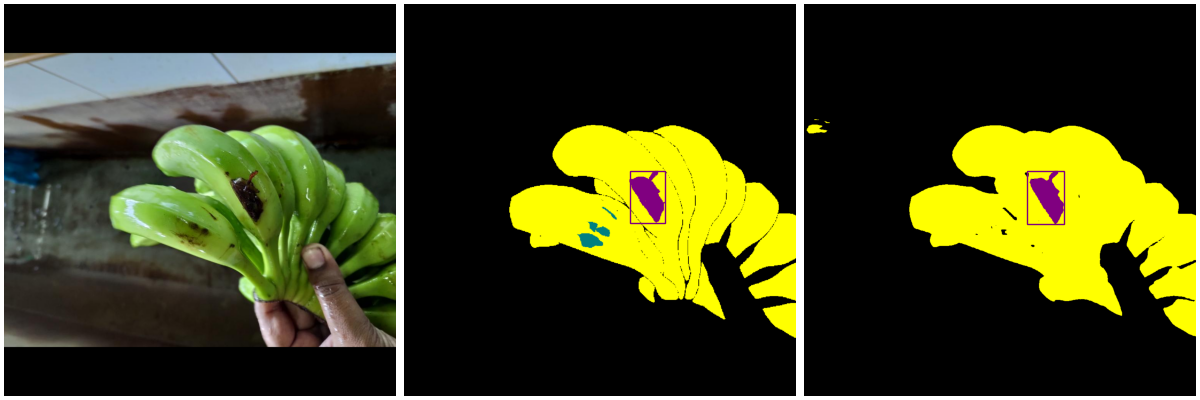}
    \includegraphics[width=.9\linewidth]{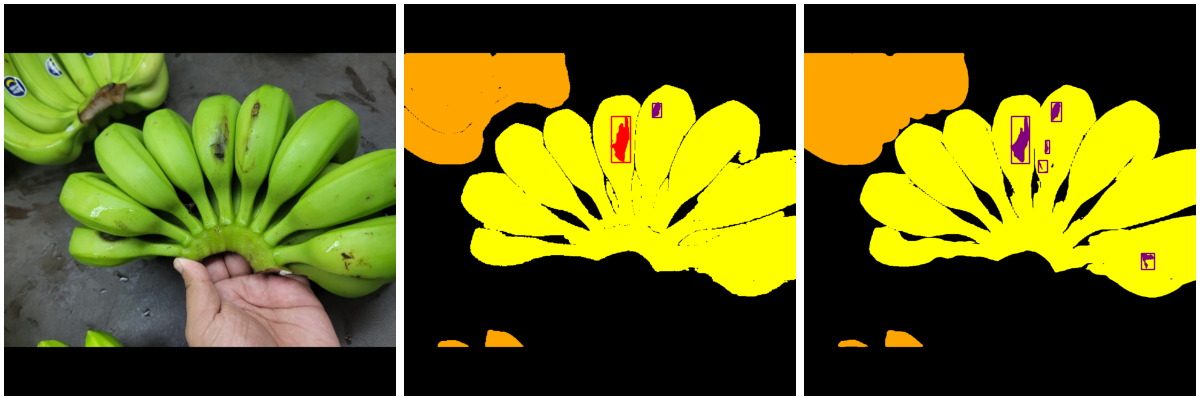}
   \caption{\textbf{Example visualizations of annotated vs. predicted masks using four defect categories.} Left: Input Image, Mid: Annotation, Right: Maskformer Prediction. Segments are color-coded as follows: \colorbox{Yellow}{Foreground Banana}, \colorbox{Orange}{Background Banana}, \colorbox{Red}{Old Bruise}, \colorbox{Purple}{Old Scar}, \colorbox{Pink}{New Bruise}, \colorbox{Teal}{New Scar}.}
    \label{fig:examples-multi}
\end{figure*}

\end{document}